\documentclass[10pt,twocolumn,letterpaper]{article}

\usepackage{iccv}
\usepackage{times}
\usepackage{epsfig}
\usepackage{graphicx}
\usepackage{amsmath}
\usepackage{amssymb}
\usepackage{multirow}
\usepackage{lipsum}
\usepackage{adjustbox}
\usepackage{booktabs}
\usepackage{makecell}
\usepackage{tabu}
\usepackage{bm}

\usepackage{xcolor}
\usepackage{colortbl}
\usepackage{color}

\usepackage[pagebackref=true,breaklinks=true,letterpaper=true,colorlinks,bookmarks=false]{hyperref}

\usepackage{cleveref}
\iccvfinalcopy % *** Uncomment this line for the final submission

\ificcvfinal\fi

\newcommand{\paragrapha}[2][1pt]{\vspace{#1}\noindent\textbf{#2}}

\definecolor{Gray}{gray}{0.95}
\definecolor{Grayy}{gray}{0.6}

\newcommand\cb[1]{\color{blue} #1}

\usepackage{soul}
\sethlcolor{Gray}
\newcommand{\PreserveBackslash}[1]{\let\temp=\\#1\let\\=\temp}
\newcolumntype{C}[1]{>{\PreserveBackslash\centering}p{#1}}
\newcolumntype{L}[1]{>{\PreserveBackslash\raggedright}p{#1}}
\ificcvfinal\fi

\usepackage{pifont}% http://ctan.org/pkg/pifont
\newcommand{\cmark}{\ding{51}}%
\newcommand{\xmark}{\ding{55}}%

\begin{document}

%%%%%%%%% TITLE
% \title{Text-to-Image Diffusion Models are Fast Learners for Visual Perception}

\title{HandMIM: Pose-Aware Self-Supervised Learning for 3D Hand Mesh Estimation}

\author{
\and
Second Author\\
Institution2\\
First line of institution2 address\\
{\tt\small secondauthor@i2.org}
}

% \author{
% %
% Wenliang Zhao$^{1}$\thanks{Equal contribution. ~~\textsuperscript{\dag}Corresponding authors.}
% ~~
% %\And
% Yongming Rao$^{1}$\samethanks
% ~~~
% %\And
% Zuyan Liu$^{1}$\samethanks
% %\And
% \\
% Benlin Liu$^{2}$
% ~~
% Jie Zhou$^{1}$
% ~~
% Jiwen Lu$^{1\dagger}$
% \\ 
% $^{1}$Tsinghua University ~~ $^{2}$University of Washington
% }

\def\spaces{~~~~~~}
\author{Zuyan Liu\thanks{Equal contribution. ~~\textsuperscript{\dag}Corresponding authors.}\spaces{}Gaojie Lin\footnotemark[1]\spaces{}Congyi Wang$^{\dagger}$\spaces{}Min Zheng\spaces{}Feida Zhu$^{\dagger}$\\\\
ByteDance}

\maketitle
\ificcvfinal\thispagestyle{empty}\fi

%%%%%%%%% ABSTRACT
\begin{abstract}
With an enormous number of hand images generated over time, unleashing pose knowledge from unlabeled images for supervised hand mesh estimation is an emerging yet challenging topic. To alleviate this issue, semi-supervised and self-supervised approaches have been proposed, but they are limited by the reliance on detection models or conventional ResNet backbones. In this paper, inspired by the rapid progress of Masked Image Modeling (MIM) in visual classification tasks, we propose a novel self-supervised pre-training strategy for regressing 3D hand mesh parameters. Our approach involves a unified and multi-granularity strategy that includes a pseudo keypoint alignment module in the teacher-student framework for learning pose-aware semantic class tokens. For patch tokens with detailed locality, we adopt a self-distillation manner between teacher and student network based on MIM pre-training. To better fit low-level regression tasks, we incorporate pixel reconstruction tasks for multi-level representation learning. Additionally, we design a strong pose estimation baseline using a simple vanilla vision Transformer (ViT) as the backbone and attach a PyMAF head after tokens for regression. Extensive experiments demonstrate that our proposed approach, named HandMIM, achieves strong performance on various hand mesh estimation tasks. Notably, HandMIM outperforms specially optimized architectures, achieving 6.29mm and 8.00mm PAVPE (Vertex-Point-Error) on challenging FreiHAND and HO3Dv2 test sets, respectively, establishing new state-of-the-art records on 3D hand mesh estimation.

\end{abstract}

\begin{figure}[t]
  \centering
   \includegraphics[width=0.9\linewidth]{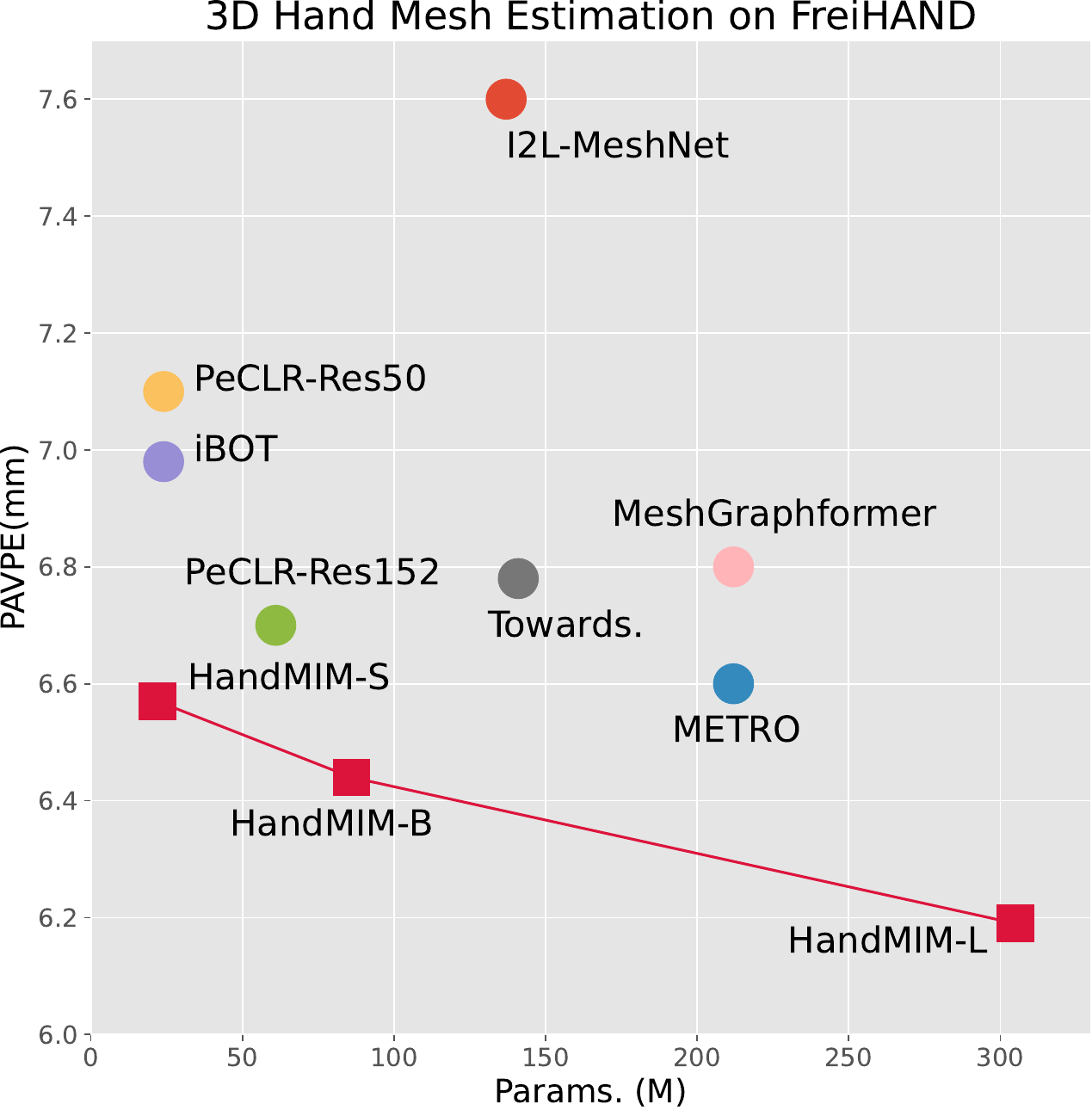}
   \caption{Performance-Paremeters trade-off of mainstream 3D hand mesh estimation methods on FreiHAND~\cite{Freihand2019} test set. We perform vertex-point-error after Procrustes alignment (lower is better). Our proposed HandMIM achieves better trade-offs in a variety of model sizes under the vanilla ViT backbone. }
   \label{fig:insight}
   \vspace{-15pt}
\end{figure}

%%%%%%%%% BODY TEXT
\section{Introduction}

Recently, 3D hand mesh estimation from monocular RGB images has drawn great attention in computer vision research~\cite{lin2021mesh, hampali2021handsformer, hampali2022keypoint, zeng2022not} driven by its potential in various applications, such as action recognition, digital human modeling, and AR/VR. However, training a high-quality hand estimation model is challenging due to complex backgrounds and severe self-occlusion. Furthermore, it is laborious and consuming to collect high-quality training pairs, especially in the format of 3D mesh. A limited amount of image-mesh training data is available, making it difficult to train effective and generalizable models. Weakly supervised methods detecting 2D keypoints~\cite{cai20203d,cai2018weakly,kulon2020weakly} or kinematic priors~\cite{spurr2020weakly} from off-the-shelf models, have been proposed to improve the accuracy of supervised-trained models. However, these methods heavily rely on detectors, such as Openpose~\cite{cao2017realtime} and MediaPipe~\cite{lugaresi2019mediapipe}, which struggle with the vast variety of wild images encountered in practice.

Self-supervised learning is a promising technique for addressing the above problem by exploiting the large quantity of unlabeled image data generated over time. Masked image modeling (MIM) pre-training has emerged as a new paradigm in self-supervised learning, which is based on the vision Transformer~\cite{dosovitskiy2020image} architecture that divides images into individual patches. In MIM pre-training, we randomly mask a specified ratio of image patches and set the self-supervised learning target as reconstructing the masked patches. Previous works~\cite{zhou2021ibot,bao2021beit,he2022masked,wei2022masked,dong2021peco,xie2022simmim} have demonstrated that MIM-based methods can learn better local and global representation than conventional self-supervised methods based on contrastive learning~\cite{oord2018representation,chen2020simple}. However, most existing self-supervised works focus on recognition tasks and aim to learn features appropriate for high-level image classification. In low-level regression tasks, mainstream methods cannot capture the equivalence of geometric transformation, which is a critical characteristic of human/hand pose or mesh. Therefore, most state-of-the-art MIM self-supervised pre-training approaches are not suitable for regression tasks such as 3D hand estimation. We confirm this finding through experiments in \cref{sec:analysis}.

In this paper, we present the first attempt to apply the effective Masked Image Modeling (MIM) self-supervised technique to 3D hand estimation tasks. We propose HandMIM, a unified and multi-granularity self-supervised pre-training strategy specially optimized for pose regression tasks. During the pre-training period, we use a teacher-student self-distillation approach, where input hand images are augmented into two views different in sizes, rotations, colors, and other factors. The student network is then tasked with reconstructing masked tokens under the guidance of the teacher network. To ensure that the class tokens are semantic with pose-aware knowledge, we introduce the pseudo keypoint alignment operation in the latent feature space. This operation allows us to undo the geometric transformation in the format of 2D keypoints, enabling the network to learn pose equivalence between cross-view tokens. To facilitate high-level and low-level recognition, we adopt token-level recovery between parallel-view tokens and pixel-level reconstruction between input images and recovered images, respectively. It is important to note that the token recovery is conducted in the \emph{same} latent space as the pose-aware alignment. 

In the supervised fine-tuning period, most existing pose estimation methods rely on the combination of grid convolution, transformer structure, and dedicated prediction head for better results. We design a simple yet strong pose estimation pipeline with a vanilla vision Transformer as the backbone, attached by a PyMAF~\cite{pymaf2021} decoder head to promote the mesh-image alignment and use the MANO~\cite{romero2022embodied} parameters to express the estimated hand mesh. We load the self-supervised pre-trained weights to transformer blocks and fine-tune the whole network for hand pose estimation. 

Extensive experiments demonstrate that our HandMIM can learn better features to improve 3D hand pose estimation precision compared with alternative self-supervised and fully-supervised methods under the same amount of labeled training data. We conduct our main experiments on two mainstream and challenging 3D hand mesh estimation datasets FreiHAND~\cite{Freihand2019} and HO3Dv2~\cite{hampali2020honnotate}. We implement HandMIM on three different sizes of vision Transformers, namely ViT-Small, ViT-Base, and ViT-Large respectively, which show strong scalability. After HandMIM pre-training, we achieve a performance boost of $9.7\%/11.9\%/16.5\%$ in PAJPE on the FreiHAND~\cite{Freihand2019} test set and $7.1\%/8.4\%/9.0\%$ on the HO3Dv2~\cite{hampali2020honnotate} test set. Notably, after pre-training HandMIM on ViT-Large, we outperform state-of-the-art methods by a large margin and establish new records on 3D hand pose estimation through the simple vanilla vision Transformer architecture. 

\noindent \textbf{Hand Pose Estimation }aims to predict hand information from a monocular RGB/depth image, and can be roughly divided into parametric and non-parametric methods. Parameter-based methods~\cite{baek2019pushing, baek2020weakly, boukhayma20193d, hasson2020leveraging, hasson2019learning, zhang2019end} use statistical priors from parametric hand models like MANO~\cite{romero2022embodied} to constrain the regression space. Except for fully-supervised manners, pioneer works~\cite{baek2019pushing, boukhayma20193d, spurr2020weakly} predict the MANO parameters with weak supervision such as hand masks and 2D annotations. Non-parametric-based methods~\cite{ge20193d, choi2020pose2mesh, kulon2020weakly} aim to predict the whole mesh vertices directly. More recent work has focused on hand-hand~\cite{moon2020interhand2, li2022interacting, fan2021learning} and hand-object interactions~\cite{yang2021cpf, Park_2022_CVPR_HandOccNet, tse2022s} which pose new and more complex challenges. In this paper, instead of designing dedicated and resource-intensive heads, we propose a lightweight head that regresses MANO parameters from a pre-trained vanilla ViT for both single-hand estimation and hand-object interaction predictions.

\begin{figure*}[t]
\begin{center}
\includegraphics[width=\linewidth]{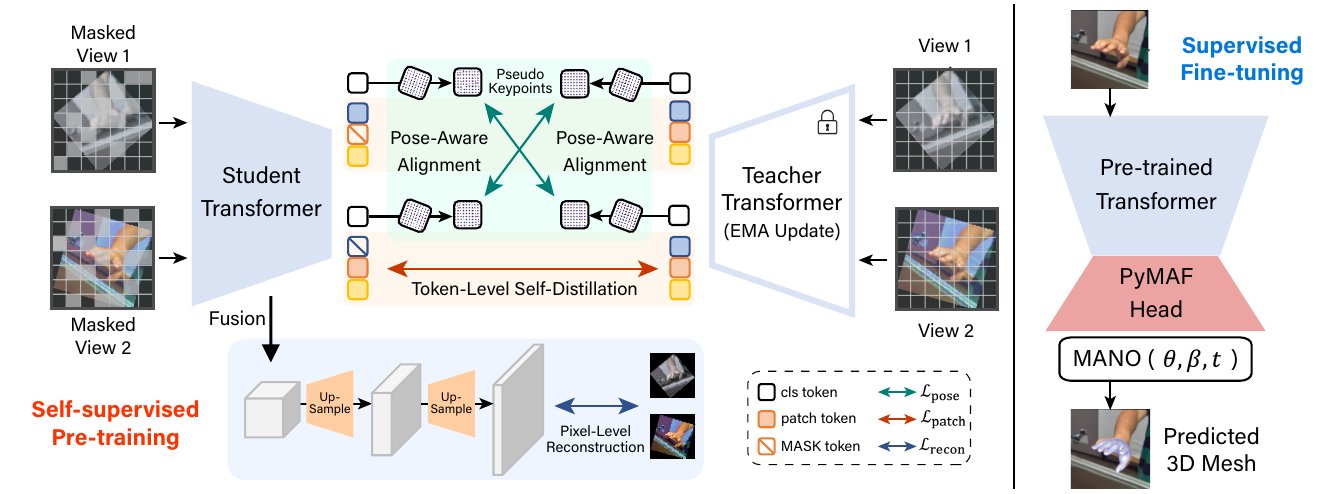}
\end{center}
   \caption{\textbf{The overall framework of HandMIM. }During the self-supervised pre-training phase, we design multi-granularity tasks to acquire pose-aware knowledge, high-level token recovery, and low-level pixel reconstruction. In the fine-tuning period, we propose a simple baseline based on the vanilla vision Transformer architecture and PyMAF~\cite{pymaf2021} decoder. We loaded the pre-trained weights onto the model, which allowed us to estimate 3D hand poses by leveraging the network's ability to learn pose-sensitive information in hand images. }
\label{fig:pipeline}\vspace{-10pt}
\end{figure*}

\vspace{1mm}
\noindent \textbf{Vision Transformers. }ViT~\cite{dosovitskiy2020image} firstly introduces vision Transformers to the visual field by patching images for transformer blocks. This approach has led to significant progress in image recognition~\cite{chen2022mobile, rao2021dynamicvit, pan2022edgevits, zeng2022not, mehta2021mobilevit} and has also shown promising results in human and hand estimation tasks~\cite{zhang2019end, lin2021mesh, hampali2021handsformer, hampali2022keypoint, zeng2022not}. For example, Mesh Graphormer~\cite{lin2021mesh} designs a transformer-based head fused with graph convolution layers. %Keypoint Transformer~\cite{hampali2022keypoint} collects candidate 2D keypoints and utilizes a transformer encoder-decoder for predictions.
However, CNN-based architecture plays a key role in geometry knowledge, therefore vanilla transformers cannot achieve competitive performance. Our approach attempts to surpass existing methods based on the vanilla ViT backbone, demonstrating the effectiveness of our pre-training algorithm.

\vspace{1mm}
\noindent \textbf{Self-supervised Learning }is an approach to learning effective feature representation from abundant unlabeled images. Contrastive learning techniques~\cite{chen2020simple,chen2020big,chen2020improved,oord2018representation,tian2020contrastive} aim to learn by constraining positive pairs to become close in feature space while pushing negative pairs apart, which have been employed in the hand pose estimation tasks for improved performance~\cite{zimmermann2021contrastive, gao2022cyclehand, spurr2021self, ziani2022tempclr}. Masked Image Modeling (MIM)~\cite{zhou2021ibot,bao2021beit,he2022masked,wei2022masked,dong2021peco,xie2022simmim} methods are new paradigms of self-supervised learning which randomly mask a portion of the input image and reconstruct the masked parts via the reasoning of other unmasked parts. The knowledge of masked images can be learned in alterable manners, including dVAE codebooks in BeiT~\cite{bao2021beit}, raw RGB pixels in SimMIM~\cite{xie2022simmim} and MAE~\cite{he2022masked}, HOG features in MaskFeat~\cite{wei2022masked}, etc. While previous MIM studies have focused on learning representative features for classification tasks, they have neglected the specificity of pose regression tasks. To our knowledge, this is the first attempt that MIM techniques have been extended to regression tasks.

\section{Method}
\label{sec:method}

In this section, we will discuss the detailed architecture of HandMIM. The pipeline of HandMIM can be found in \cref{fig:pipeline}. We start with preliminaries including basic knowledge of vision transformers and masked image modeling in \cref{sec:preliminaries}. Then we introduce the detailed design of HandMIM, including pose-aware keypoint alignment in \cref{sec:pose}, token-level self-distillation in \cref{sec:token}, and pixel-level reconstruction in \cref{sec:pixel}. Finally, we illustrate how to apply pre-trained features after self-supervised learning for 3D hand mesh estimation tasks in \cref{sec:mesh}. 

\subsection{Preliminaries}
\label{sec:preliminaries}

\paragrapha{Vision Transformers. }Given input images $x=\{x_i\}_{i=1}^N\in \mathbb{R}^{3\times H\times W}$, vision transformer~\cite{dosovitskiy2020image} applies a patch embedding layer to divide the images into patch tokens $z\in \mathbb{R}^{n^2\times c}$, where $n$ is determined by a pre-defined patch length. A class token is appended to the patch tokens to generate the input feature $z\in \mathbb{R}^{n^2+1\times c}$ of transformer blocks. Each transformer block consists of a multi-head self-attention (MHSA) and feed-forward network (FFN). The input tokens are projected into $\mathbf{Q},\mathbf{K},\mathbf{V}$ through linear layers, and the forward process of MHSA can be formulated as follows:
\begin{equation}
    \text{MHSA}(\mathbf{Q},\mathbf{K},\mathbf{V})=\text{Softmax}\left(\dfrac{\mathbf{Q}\mathbf{K}^T}{\sqrt{d}}\right)\mathbf{V}
\end{equation}
The output of the self-attention module is then passed through an inverted bottleneck MLP, also known as the feed-forward network. In practice, vision Transformers are assembled by stacking a series of transformer blocks. By varying the channel width and layer depth of vision Transformers, we can obtain models of varying sizes. 

\paragrapha{Masked Image Modeling. }Masked Image Modeling (MIM) is a self-supervised learning technique that has been demonstrated to be a general method for image recognition tasks in a number of recent works~\cite{xie2022simmim,bao2021beit}. Given input tokens $z\in \mathbb{R}^{n^2\times c}$, and then randomly create a binary mask $\mathcal{M}\in \{0,1\}^{n\times n}$. When $\mathcal{M}=1$, the origin image tokens are passed through the neural network backbone, and when $\mathcal{M}=0$, the input tokens are replaced with a special mask token $p_{mask}$. By doing so, we obtain both the original tokens $z_i$ and the masked tokens $\hat{z}_i$, which are calculated as  $\hat{z}_i=\mathcal{M}\odot p_{mask}+(1-\mathcal{M})\odot z_i$. The goal of the masked image modeling task is to train the backbone function $f(\cdot)$ to recognize and recover the original tokens $z_i$ from the masked tokens $\hat{z}_i$:
\begin{equation}
    \mathcal{L}=\sum_{i=1}^N\mathcal{M}_i\cdot  \|f(\hat{z}_i) - z_i  \|_2
\end{equation}

\subsection{Pose-Aware Keypoint Alignment}
\label{sec:pose}

We observe that the pose of hands in input images remains equivalent after data augmentation, such as random rotation and resizing operations, while the positional information is altered. Existing mainstream self-supervised learning methods fail to capture the knowledge of poses. In HandMIM, we propose the idea of pose-aware keypoint alignment to extract pose-relevant knowledge. 
Consider a point $P=(x_0,y_0)$ in input image $x_i$, after the augmentation process, the point is transformed to: 
\begin{equation}
(x_0',y_0')=R((x_0,y_0)\cdot \gamma - (a,b))    
\end{equation}
\noindent where $R$ denotes 2D rotation matrix, $\gamma$ denotes scale factor and $(a,b)$ denotes upper left coordinate of the resized image. After the transformer backbone, we obtain the output class token $\tau\in \mathbb{R}^{1\times c}$ in latent space, where we can regard it as a set of \textit{pseudo points} and reshape $\tau$ into the format of point $(\tau_x, \tau_y)\in \mathbb{R}^{1\times c/2}$. We can then recover the transformation to get the original latent feature before augmentation as follows: 
\begin{equation}
    (\tau_x', \tau_y')=1/\gamma\cdot(R^{-1}(\tau_x, \tau_y)+(a,b))
\end{equation}

After keypoint alignment in latent space, image features after different augmentations exhibit a unified hand pose, facilitating the extraction of pose-sensitive knowledge by vision backbones. We will elaborate on how to learn the pose-aware task in the following subsection. 

\subsection{Token-Level Self-Distillation}
\label{sec:token}

The knowledge of masked image modeling can be acquired through a self-distillation approach proposed by DINO~\cite{caron2021emerging}. We treat self-supervised learning as a discriminative task involving two backbones with identical architecture, which play the roles of a teacher network $P_t$ and a student network $P_s$. Specifically, we train the student network to comprehend corrupted input tokens $\hat{z}_i$ under the guidance of the teacher network, which receives complete input tokens $z_i$.

To fully recognize the images, we use two random augmentations, denoted as $\mu$ and $\nu$, thus we get augmented tokens $u_i=\mu(z_i), v_i=\nu(z_i)$ for the teacher network $f_t$. We then apply a randomly generated mask $\mathcal{M}_i$ to the augmented tokens after the patch embedding layer, resulting in corrupted tokens $\hat{u}_i$ and $\hat{v}_i$ for the student network $f_s$. The process in the student and teacher networks can be formulated as follows:
\begin{align}
 \label{eq:method1} \hat{U}_i=f_s(\hat{u}_i), \hat{V}_i=f_s(\hat{v}_i), U_i=f_t(u_i), V_i=f_t(v_i)
\end{align}

We use uppercase letters, i.e., $U$ and $V$, to denote the output of the backbones $f$. The class tokens, denoted by $U_{[cls]}$ and $V_{[cls]}$, contain high-level semantic knowledge, while the patch tokens, denoted by $U_{[patch]}$ and $V_{[patch]}$, contain low-level local knowledge of the input images.

We design specific tasks of self-supervised learning for class considering their semantic meanings. For the class tokens, we aim to extract the pose of the original images, which is equivalent after the inverse operation of spatial data augmentations, implemented as pose-aware keypoint alignment in \cref{sec:pose}. Since we expect images under different augmentations to have the same pose expression, we adopt a cross-entropy loss between the \emph{cross-view} images and apply the self-distillation approach to enable learning from the teacher to the student network. Specifically, we obtain the $\mathcal{L}_\text{pose}$ loss, which can be formulated as follows:
\begin{equation}
    \mathcal{L}_\text{pose}=-U_{i[{cls}]}\log \hat{V}_{i[{cls}]}-V_{i[{cls}]}\log \hat{U}_{i[{cls}]}
    \label{eq:method2}
\end{equation}

During the backward period, only the student network requires a gradient backpropagation as we treat the output of the teacher network as ground truth. Subsequently, we update the teacher network through Exponentially Moving Averaged (EMA) using the student network.

Given the patch output of the transformer backbone which represents the spatial knowledge of input images, we can define the patch loss $\mathcal{L}_{\text{patch}}$. This loss measures the discrepancy between the \emph{parallel view} tokens, which share the same spatial position after the augmentations. Specifically, we aim to train our module to recover the corrupted patch tokens. We learn the knowledge using a similar self-distillation approach as in \cref{eq:method2}:
\begin{equation}
    \mathcal{L}_{\text{patch}}=-U_{i[{patch}]}\log \hat{U}_{i[{patch}]}-V_{i[{patch}]}\log \hat{V}_{i[{patch}]}
\end{equation}

\subsection{Pixel-Level Reconstruction}
\label{sec:pixel}

Hand pose estimation is a low-level task that involves directly analyzing image pixels, in contrast to image classification. While token-level self-distillation may be effective for higher-level knowledge, it may lack the necessary low-level understanding. To address this, we propose a pixel-level reconstruction module. Since transformer tokens are applied in a patch-based manner, we integrate a pyramid fusion layer following certain transformer blocks and gradually up-sample using transposed convolution (T-Conv). The resulting pyramid fusion output ($T^i$) is concatenated with the transformer block output ($L^i$) and fused using a linear layer. Mathematically, this can be represented as follows:
\begin{equation}
    T^{i+1}=\text{T-Conv}(\text{Linear}(\text{Concat}[T^i, L^i]))
\end{equation}
In common practice, vision transformers use a patch size of 16, therefore 4 times of transposed convolution are adopted to recover the original shape of input $x$, we can adopt L1-Loss between input images and reconstruction results:
\begin{equation}
    \mathcal{L}_{\text{recon}}=\mathcal{M}\odot \|(T^{4}-x)\|_{1}
\end{equation}
where $\mathcal{M}$ denotes the token mask and $\odot$ denotes the Hadamard product. Note that only the student network requires gradient, therefore we only adopt $\mathcal{L}_{\text{Recon}}$ at the student network with masked input. 

The final loss function is the sum of the losses mentioned above:
\begin{equation}
    \mathcal{L}=\mathcal{L}_{\text{pose}}+\mathcal{L}_{\text{patch}}+\mathcal{L}_{\text{recon}}
\end{equation}

\subsection{3D Hand Mesh Estimation via ViT}
\label{sec:mesh}

To evaluate the effectiveness and benefits of HandMIM self-supervised pre-training, we fine-tune the pre-trained vision transformer backbone on a supervised 3D hand mesh estimation task. Specifically, we incorporate a keypoint feedback loop after the backbone, similar to the approach used in PyMAF~\cite{pymaf2021}, to predict MANO~\cite{romero2022embodied} parameters, including joint rotation ($\theta$), shape coefficient ($\beta$), and global translation ($t$). This keypoint feedback loop is composed of three rectifier layers that extract local features based on the current keypoint-image alignment status and feed them back for rectification.

To train our method, we use a combination of MANO parameter loss ($\mathcal{L}_{\text{MANO}}$), vertice loss ($\mathcal{L}_{\text{vert}}$), and keypoint loss ($\mathcal{L}_{\text{kpt}}$), which are described as follows.
\begin{equation}
    \mathcal{L_\text{FT}}= \mathcal{L}_{\text{MANO}} +\mathcal{L}_{\text{vert}} +  \mathcal{L}_{\text{kpt}} 
\end{equation}

The MANO parameter loss ($\mathcal{L}_{\text{MANO}}$) is calculated as the L2 distance between the predicted MANO parameters and the ground truth. Given MANO parameters ($\theta$, $\beta$, $t$), the 3D mesh vertices can be obtained using the MANO model $M(\theta,\beta)\rightarrow\mathbb{R}^{778}$, which can be used to calculate the vertice loss ($\mathcal{L}_{\text{vert}}$) as a more direct form of supervision. Furthermore, the 3D keypoints $J_{3D}\in\mathbb{R}^{21\times3}$ can be generated by mapping 3D mesh using a pre-trained linear regressor. By projecting the 3D keypoints $J_{3D}$ onto the image coordinate system, we can obtain 2D keypoints, which can be used to supervise the training process with 2D keypoint ground truth $\hat{J}_{2D}$. Overall, the keypoint loss $\mathcal{L}_{\text{kpt}}$ is composed of 3D keypoint loss and projected 2D keypoint loss as follows:
\begin{equation}
    \mathcal{L}_{\text{kpt}}= \|J_{3D} - \hat{J}_{3D}\|_{1} +  \|(K(J_{3D}+t) - \hat{J}_{2D}\|_{1}
\end{equation}
where $K$ indicates the ground-truth camera intrinsic matrix following common practice.

\section{Experiments}
\label{sec:exp}

In this section, we conduct extensive experiments to evaluate the proposed self-supervised keypoint pre-training framework HandMIM. We first introduce our strategy on HandMIM pre-training on \cref{sec:exp-pretrain}. Then we show the results of our pre-trained model on 3D hand mesh estimation tasks on \cref{sec:exp-mesh}. Finally, we present in-depth analysis and ablation studies on \cref{sec:analysis}. 

\subsection{HandMIM Pre-training}
\label{sec:exp-pretrain}

%%%%%%%%%%%%%%%%%%%%%%%% table sota FH %%%%%%%%%%%%%%%%%%%%%%%
\renewcommand\arraystretch{1.1}
\begin{table}
\centering    
\caption{\small \textbf{Results on FreiHAND~\cite{Freihand2019} dataset. }We perform our results before and after HandMIM pre-training and list the lifting ratio compared with ViT-S baseline. All the results use the same labeled dataset for supervised learning. $*$ denotes non-ensemble evaluation results for a fair comparison. HandMIM outperforms existing methods by a large margin. }\vspace{3pt}
 \adjustbox{width=\linewidth}{
 \setlength{\tabcolsep}{2.5pt}
	\begin{tabular}{L{120pt}L{50pt}L{45pt}C{35pt}C{35pt}}\toprule
		Methods  & PAVPE$\downarrow$ & PAJPE$\downarrow$ & F@5$\uparrow$ & F@15$\uparrow$  \\ \midrule
		I2L-MestNet~\cite{Moon_2020_ECCV_I2L-MeshNet}  &7.6  &7.4 &0.681  &0.973 \\
		I2UV-HandNet\cite{chen2021i2uv}  &7.4  &7.2  &0.707  &0.977 \\
		HIU-DMTL~\cite{zhang2021hand}  &7.3  &7.1  &0.699 &0.974  \\
		Tang et al.~\cite{tang2021towards}  &7.1 &7.1 &0.706 &0.977  \\
		PeCLR~\cite{spurr2021self}  &- &7.1 &- &-  \\
		Mesh Graphormer$^*$ ~\cite{lin2021mesh} &6.8 &6.6  &0.732 &0.982  \\
		  METRO$^*$ ~\cite{lin2021end} &6.6 &6.3 &0.715 &0.981  \\
		MobRecon~\cite{chen2022mobrecon} &7.2 &6.9 &0.694 &0.979 \\
		ViT-Small &7.1 &7.2 &0.697 &0.978   \\
		ViT-Large &6.6 &6.6 &0.727 &0.983   \\
		\midrule
	    \rowcolor{Gray} HandMIM-Small & $6.57_{\cb{-8.1\%}}$ & $6.57_{\cb{-9.7\%}}$ &0.725 &0.984 \\
		% ViT-Base &6.76 &6.76 &0.715 &0.982   \\
		\rowcolor{Gray} HandMIM-Base & $6.47_{\cb{-9.7\%}}$ & $6.44_{\cb{-11.9\%}}$ &0.731 &0.985 \\
		% ViT-Large &6.58 &6.52 &0.727 &0.983   \\
		\rowcolor{Gray} HandMIM-Large & $\textbf{6.29}_{\cb{-12.9\%}}$ & $\textbf{6.19}_{\cb{-16.5\%}}$  &\textbf{0.744} &\textbf{0.986}   \\
		\bottomrule
	\end{tabular}}
\vspace{-12pt}
\label{table:sota_fh}
\end{table}

\begin{figure*}[t]
  \centering
     \hspace{-6mm} \begin{minipage}[t]{0.255\textwidth}
        \centering
        \includegraphics[height = 3.3 cm,width=4.3cm]{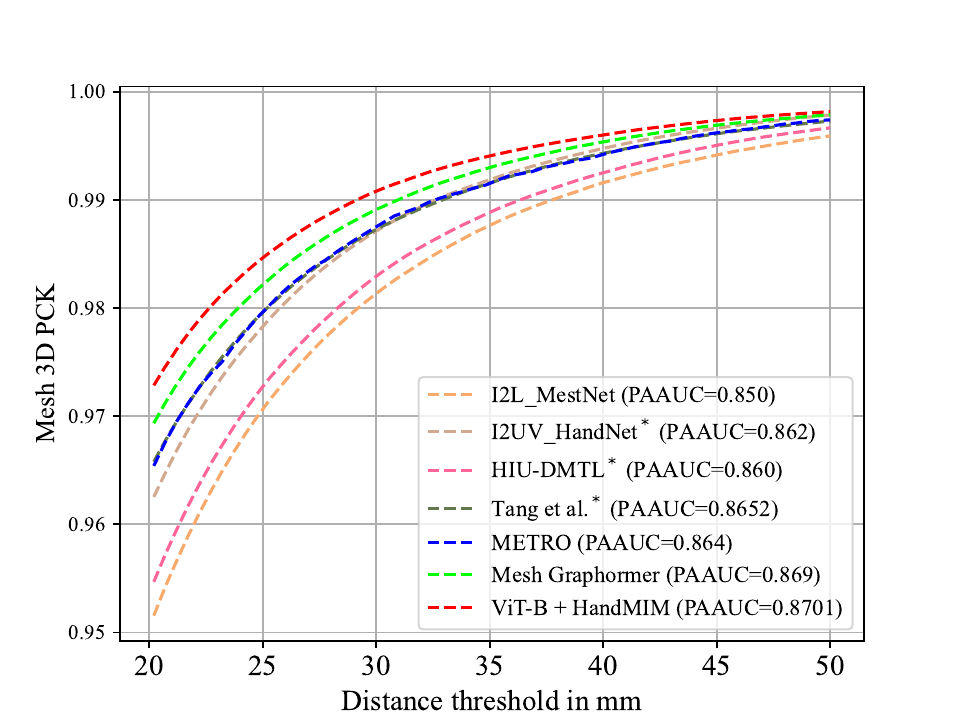}
        \centerline{\scriptsize{(a) FreiHAND Mesh AUC}}\medskip
    \end{minipage}%
    \begin{minipage}[t]{0.255\textwidth}
        \centering
        \includegraphics[height = 3.3 cm,width=4.3cm]{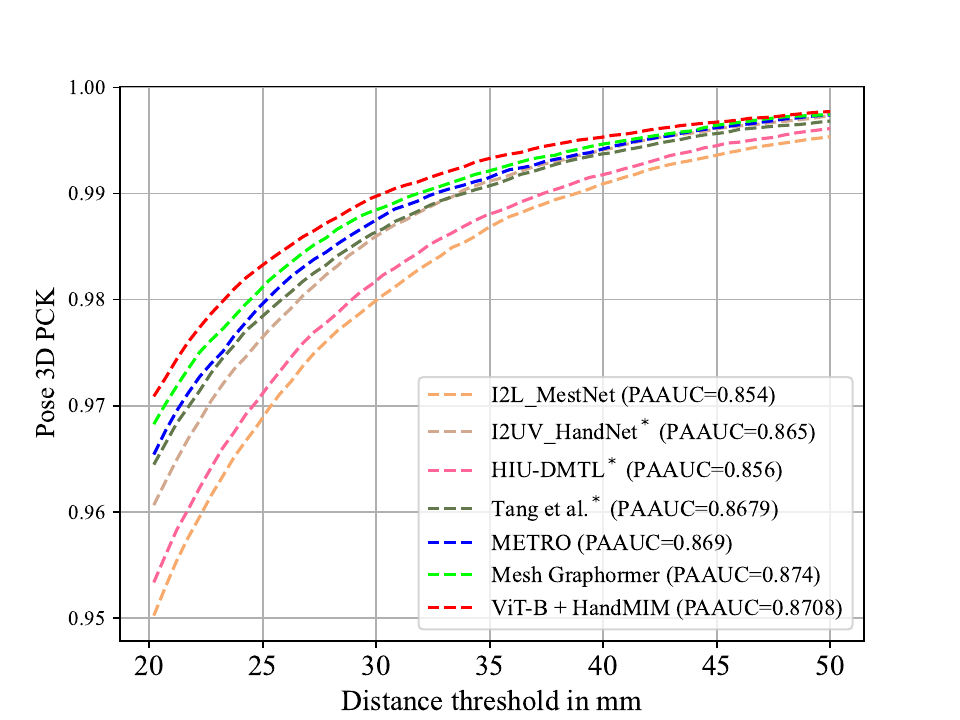}
        \centerline{\scriptsize{(b) FreiHAND Pose AUC}}\medskip
    \end{minipage}%
        \begin{minipage}[t]{0.255\textwidth}
        \centering
        \includegraphics[height = 3.3 cm,width=4.3cm]{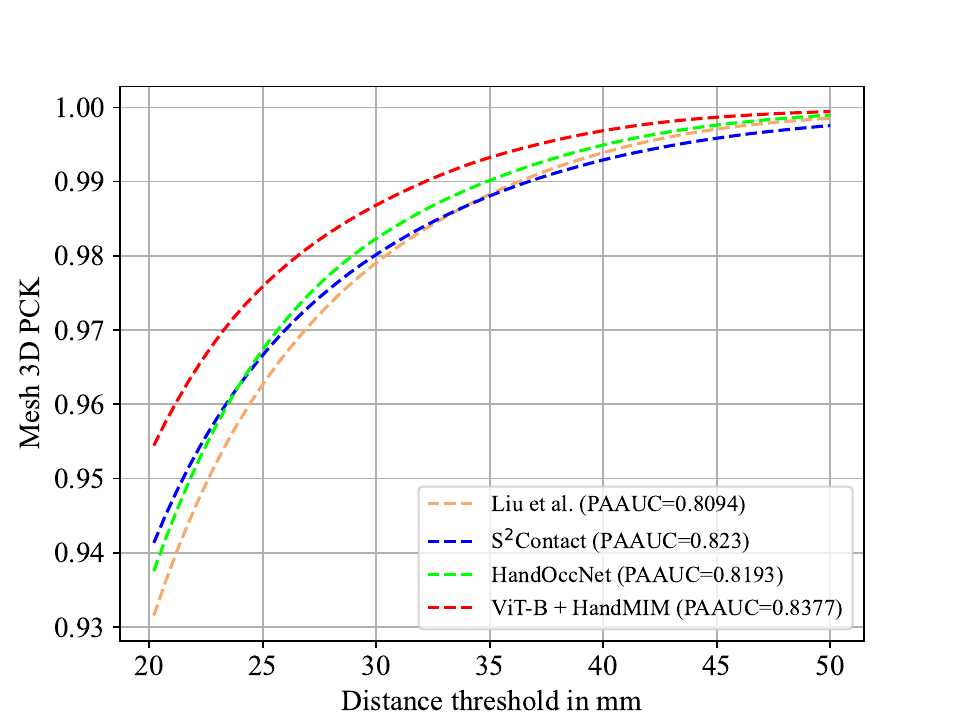}
        \centerline{\scriptsize{(c) HO3D v2 Mesh AUC}}\medskip
    \end{minipage}
    \begin{minipage}[t]{0.255\textwidth}
        \centering
        \includegraphics[height = 3.3 cm,width=4.3cm]{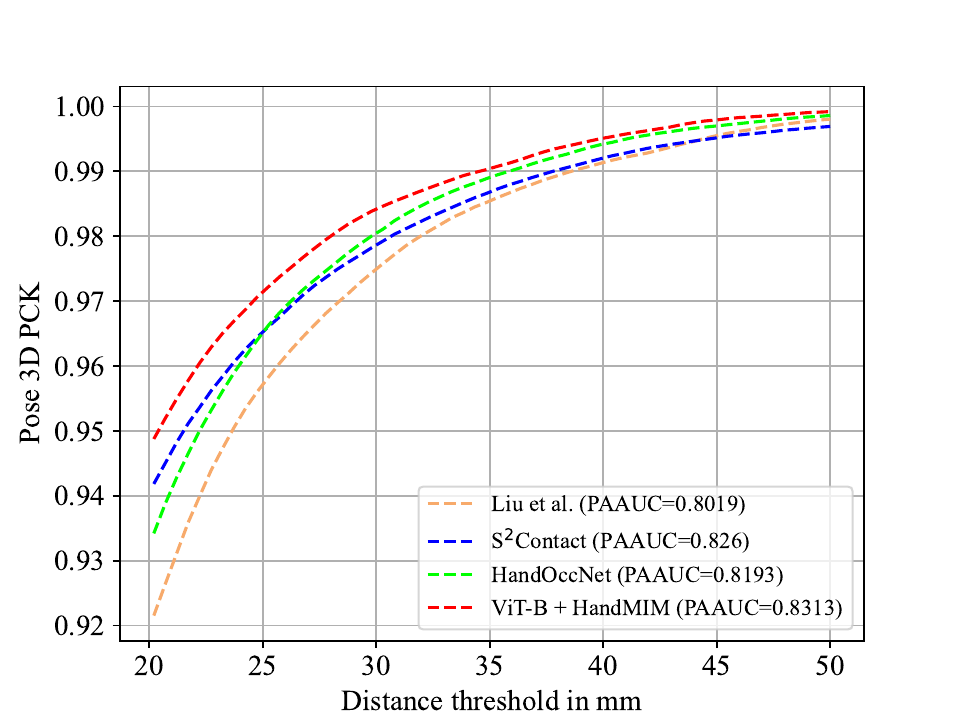}
        \centerline{\scriptsize{(d) HO3D v2 Pose AUC}}\medskip
    \end{minipage}%
    % \vspace{-2mm}
\caption{\small \textbf{The pose and mesh AUC comparison with other state-of-the-art on FreiHAND and HO3D datasets.} $^*$ indicates the method is supervised and trained with extra 2D/3D labeled data. It can be observed from the plot that our method achieves the best performance on both datasets for both mesh and pose AUC values with ViT-B as the backbone. }\label{fig:auc_curve}
\vspace{-12pt}
\end{figure*}

\noindent \textbf{Pre-training Settings. }
We employ Vision Transformers~\cite{dosovitskiy2020image} as our backbone in different sizes, including ViT-Small, ViT-Base, and ViT-Large. Details of the architectures can be found in the supplementary materials. We collect the multi-level features from layer $[3,6,9,12]$ for pixel reconstruction with a decoder consisting of linear layers for feature fusing and transposed convolutions for up-sampling. Input images are augmented through random resizing within the range $(0.08, 1)$, rotating within the range $(0, 150^{\circ})$, followed by color jitter, grayscale, gaussian blur, and solarization. After the backbone, a shared MLP is used to project the tokens into latent space. We then resize the class token into high-level pseudo keypoints with size $[128, 2]$. During HandMIM pre-training, we use AdamW~\cite{loshchilov2017decoupled} as the optimizer with a batch size of 1024. We pre-train ViT-S and ViT-B for 400 epochs, and ViT-L for 250 epochs. The learning rate is set to 2e-3, and the masked ratio $r$ is randomly sampled within the range $[0.1, 0.5]$. 

\noindent\textbf{Pre-training Datasets. }As there are currently no standardized datasets for hand pose self-supervised learning, we collect hand images across a variety of datasets for sufficient hand pose and background distributions, including FreiHAND~\cite{Freihand2019} training set, Youtube3DHands~\cite{kulon2020weakly}, COCO 2017 train and unlabeled images\cite{jin2020whole}. For datasets without hand annotations, we use MediaPipe~\cite{lugaresi2019mediapipe} to detect hands and crop the images accordingly. Our dataset consists of a total of 221,602 images for self-supervised learning.

\subsection{3D Hand Mesh Estimation}
\label{sec:exp-mesh}

We evaluate the performance of HandMIM models against several state-of-the-art methods in 3D hand mesh estimation. Our experiments demonstrate that pre-training HandMIM models significantly enhance the accuracy and quality of visualizations in 3D hand mesh estimation tasks and achieve state-of-the-art performance in multiple datasets and metrics.

\noindent\textbf{Setups. }We use two challenging publicly available hand pose estimation datasets FreiHAND\cite{Freihand2019} and HO3D v2~\cite{hampali2020honnotate} in our experiments. The FreiHAND dataset comprises 130,240 training images with a green screen or composite background and 3,960 test images with a real background. HO3Dv2 is a hand-object interaction dataset with a more complex occlusion, and its evaluation process is conducted online. During training, we set the batch size to 128, and we crop and resize the hand image to $224\times224$. Random scale, translation, rotation, and color jitter are applied for data augmentation. We fine-tune our model using the Adam optimizer \cite{kingma2014adam} for 100 epochs, with a learning rate of $4e^{-5}$. Our ViT-S model achieves a real-time inference speed of 40 FPS on a single NVIDIA V100 GPU.
\renewcommand\arraystretch{1.1}
\begin{table}
\centering
    \caption{\small \textbf{Results on HO3D v2~\cite{hampali2020honnotate} dataset. } Compared with current methods specially designed for hand-object interaction, we achieve better results under a vanilla backbone with no special operation. All the listed results use the same labeled dataset for supervised learning. }\vspace{3pt}
	\resizebox{0.49\textwidth}{!}
    {
	\begin{tabular}{lllccc}\toprule
		Methods  & PAVPE$\downarrow$ & PAJPE$\downarrow$ &MPJPE$\downarrow$ & F@5$\uparrow$ & F@15$\uparrow$  \\
		\midrule
		Liu et al.~\cite{liu2021semi}  &9.5  &9.9 &31.7 &0.528  &0.956 \\
		HandOccNet\cite{Park_2022_CVPR_HandOccNet}  &8.8  &9.1 &24.0 &0.564  &0.963 \\
		S$^2$Contact~\cite{tse2022s}  &8.86  &8.74 &-  &- &-  \\
		ArtiBoost~\cite{li2021artiboost}  &10.9  &11.4 &25.3 &0.488 &0.944  \\
		ViT-Small &8.78 &9.18 &26.37 &0.567 &0.963   \\
		ViT-Large  &8.43 &8.73 &23.57 &0.588 &0.970   \\
		\midrule
	    \rowcolor{Gray}HandMIM-Small & $8.22_{\cb{-6.8\%}}$ & $8.57_{\cb{-7.1\%}}$  &24.00 &0.597 &0.970  \\
		% ViT-Base  &8.60 &8.98 &24.80 &0.582 &0.966   \\
		\rowcolor{Gray}HandMIM-Base & $8.08_{\cb{-8.0\%}}$  & $8.41_{\cb{-8.4\%}}$ &22.01 &0.610 &0.971   \\
		\rowcolor{Gray}HandMIM-Large & $\textbf{8.00}_{\cb{-8.9\%}}$  & $\textbf{8.35}_{\cb{-9.0\%}}$   &\textbf{21.94} &\textbf{0.617} &\textbf{0.972}   \\
		\bottomrule
	\end{tabular}
	}\vspace{-12pt}
	\label{table:sota_HO3Dv2}
    
\end{table}

\noindent\textbf{Evaluation Metrics.} We incorporate multiple evaluation metrics for comprehensive analysis and comparison. We use \emph{Joint-Point-Error (JPE)} and \emph{Vertex-Point-Error (VPE)} to denote the average L2 distance between the ground truth and predicted keypoints and mesh vertices, respectively. We prefix the metrics with \emph{PA} and \emph{MP} to denote Procrustes alignment and scale-and-translation alignment. \emph{F-scores} is defined as the harmonic mean between recall and precision between two meshes given a distance threshold. We also report the \emph{Area Under Curve (AUC)} following common practice. We report our evaluation results in $mm$ unit by default.

\noindent\textbf{Results on FreiHAND. }
We compare our approach with existing methods~\cite{Moon_2020_ECCV_I2L-MeshNet, chen2021i2uv, zhang2021hand, tang2021towards, spurr2021self, lin2021end, lin2021mesh, chen2022mobrecon} on mainstream FreiHAND dataset. We conduct self-supervised pre-training with HandMIM using ViT-Small (ViT-S), ViT-Base (ViT-B), and ViT-Large (ViT-L). As shown in \cref{table:sota_fh}, fine-tuning our approach using HandMIM pre-trained weights consistently improves the performance on both datasets compared to the baseline, confirming the effectiveness of HandMIM pre-training. We plot the mesh and pose AUC in \cref{fig:auc_curve}. Notably, even with the lightweight ViT-Small with 22M parameters, our approach achieves a competitive PAVPE of \emph{6.6mm}, which further improves to the best PAVPE of \emph{6.29mm} when we employ ViT-L as the backbone.

\noindent\textbf{Results on HO3D v2. }
For HO3D v2, existing methods~\cite{liu2021semi, Park_2022_CVPR_HandOccNet, tse2022s, li2021artiboost} design various complex strategies via hand-object interaction information to improve the estimation accuracy. For example, HandOccNet~\cite{Park_2022_CVPR_HandOccNet} carefully designs a network to tackle severe hand occlusion. S$^2$Contact~\cite{liu2021semi} learns hand-object contact clues to refine inaccurate pose estimations. In contrast, our proposed HandMIM approach trains a simple and lightweight hand-to-mesh regression model that achieves superior results without relying on complex strategies. As shown in Table \ref{table:sota_HO3Dv2}, even when using a ViT-S model with only 22M parameters, HandMIM achieves a PAVPE score of \emph{8.22mm}, surpassing other existing methods by a significant margin. We also plot the mesh and pose AUC in \cref{fig:auc_curve}. This demonstrates the robustness of our model, particularly in handling severe hand-object occlusion.

%%%%%%%%%%%%%%%%%%%%%%%%%% Vis of Cmp FH %%%%%%%%%%%%%%%%%%%%%%%%%%
\begin{figure}[t]
	\centering
	\includegraphics[width = .48 \textwidth]{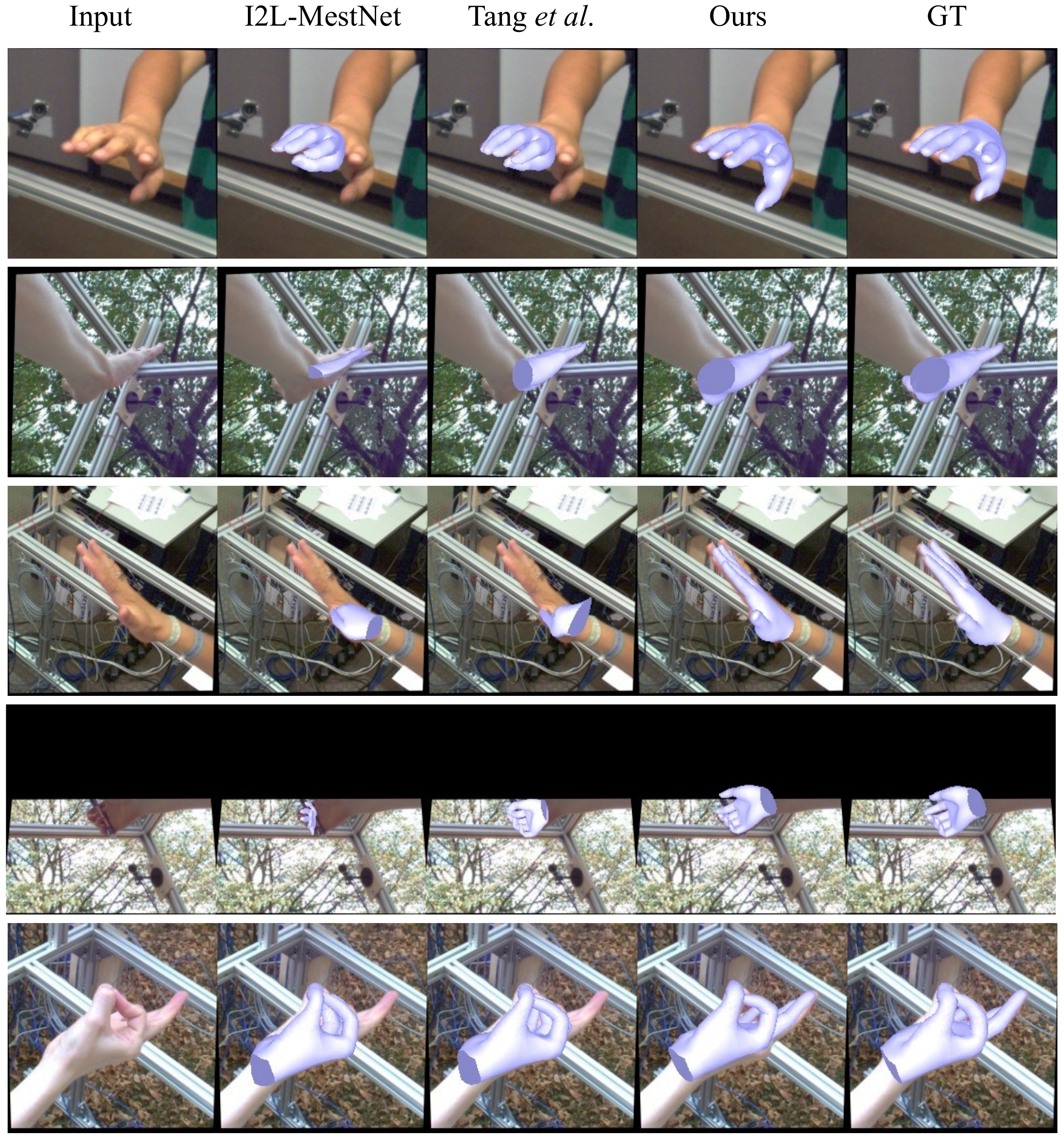}
    \caption{\small \textbf{Visualizations on FreiHAND~\cite{Freihand2019} test set. } From left to right we show the input images, the predictions from I2L-MeshNet~\cite{Moon_2020_ECCV_I2L-MeshNet}, Tang et al.~\cite{tang2021towards}, our HandMIM and the ground truth. Our method is more robust for hard view-point, occlusion, and complicated hand gestures.}
	\label{fig:cmp_vis_fh_sota}\vspace{-12pt}
\end{figure}
\begin{figure}[t]
	\centering
	\includegraphics[width = .48 \textwidth]{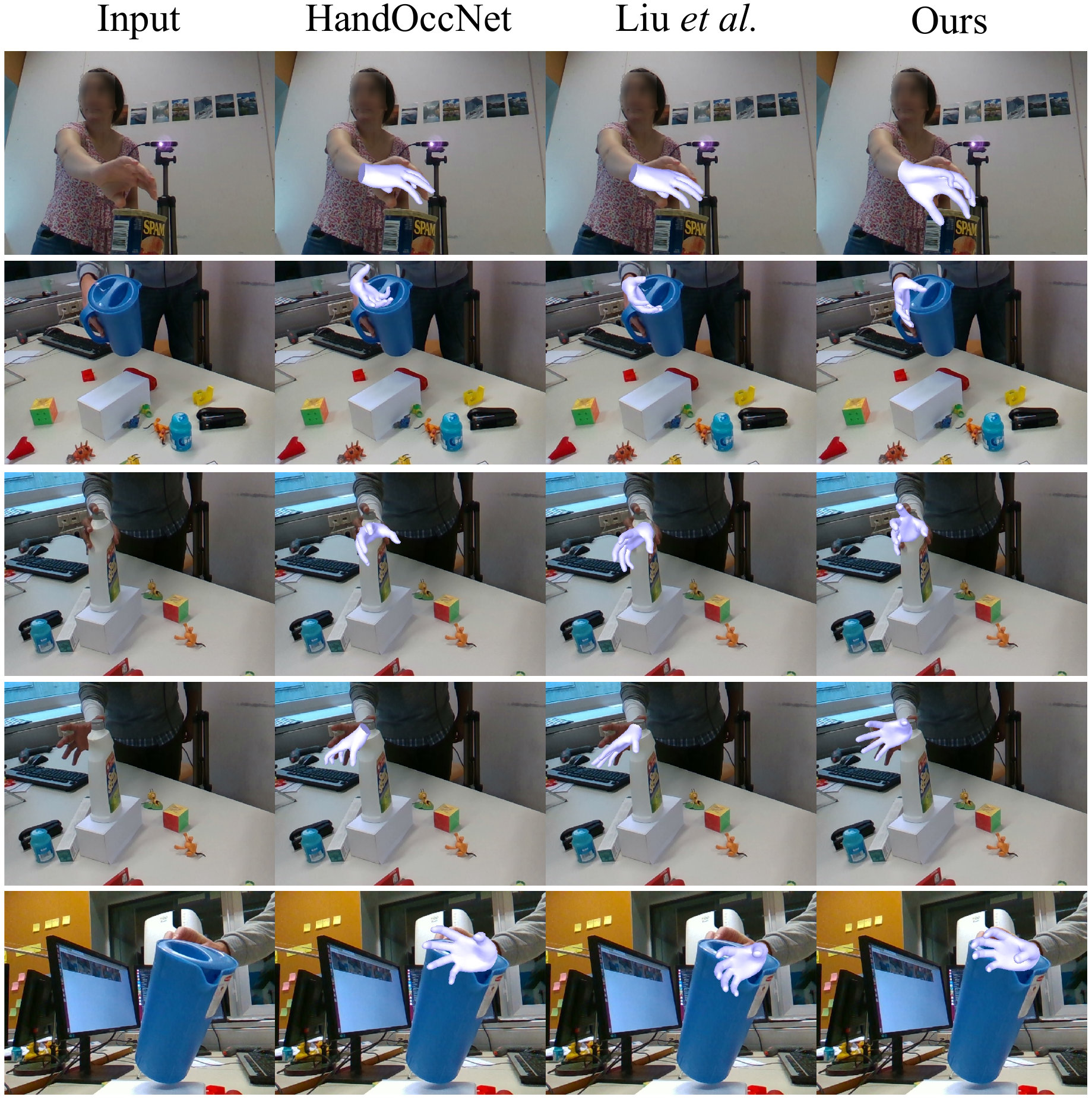}
    \caption{\small \textbf{Visualizations on HO3D v2~\cite{hampali2020honnotate} test set. } We show the input images, the predictions from HandOccNet~\cite{Park_2022_CVPR_HandOccNet}, Liu et al.~\cite{liu2021semi} and our HandMIM. We can observe that our method captures more precise poses under the corruption and occlusion of complex objects. }
	\label{fig:cmp_vis_HO3Dv2_sota}\vspace{-5pt}
\end{figure}

\noindent\textbf{Visualizations.} We visualize and compare the hand mesh predictions of our proposed method with state-of-the-art on the test sets of FreiHAND~\cite{Freihand2019} and HO3Dv2~\cite{hampali2020honnotate} in Figures \ref{fig:cmp_vis_fh_sota} and \ref{fig:cmp_vis_HO3Dv2_sota}, respectively. Our method achieves better estimation accuracy for challenging viewpoints, severe occlusion, and difficult gestures compared to existing methods. For images in the HO3Dv2 dataset under severe hand-object occlusion, our method can capture local finger clues and infer the overall wrist pose and plausible finger positions, demonstrating its superior robustness compared to alternative methods.

\subsection{Analysis}
\label{sec:analysis}

In this subsection, we present a series of convincing analysis experiments and ablations to evaluate the effectiveness of HandMIM. We demonstrate the superiority of our method against existing self-supervised methods through comprehensive comparisons. To evaluate the generalizability of our method, we perform cross-dataset, partial fine-tuning analysis, and visualizations of HandMIM.  

%%%%%%%%%%%%%%%%%%%%%%%%%% Vis of Cmp FH %%%%%%%%%%%%%%%%%%%%%%%%%%

\noindent\textbf{Comparisons with alternative self-supervised learning methods. }As shown in \cref{table:semi_cmp}, we compared the performance of our proposed pose-aware method for 3D hand mesh estimation with two representative self-supervised learning methods, the mainstream masked image modeling method iBOT~\cite{zhou2021ibot} and the contrastive-learning-based method PeCLR~\cite{spurr2021self}. We conducted these comparisons using \emph{the same} ViT-Small backbone and \emph{the same} amount of training data. Our results indicate that HandMIM outperforms iBOT, which is a representative MIM method used for visual recognition tasks. Furthermore, we observed that the contrastive-learning-based method is not suitable for stronger vision Transformer architectures, resulting in a significant accuracy drop. These findings demonstrate the superiority of our proposed method over existing self-supervised learning methods for hand estimation tasks.

\begin{table}
\centering
\caption{\small \textbf{Comparisons with self-supervised methods. } We train HandMIM together with baselines under the same backbone and the same training data. Results are evaluated on FreiHAND~\cite{Freihand2019} test set using the same regression head. PeCLR~\cite{spurr2021self} shows an accuracy drop based on stronger vision transformers. Our HandMIM outperforms existing methods by a large margin. }\vspace{3pt}
	\resizebox{0.48\textwidth}{!}
    {
	\begin{tabular}{lccccc}\toprule
		Methods  & Dataset & PAVPE$\downarrow$ & PAJPE$\downarrow$ & F@5$\uparrow$ & F@15$\uparrow$  \\\midrule
		ViT-S & FH &7.10  &7.21 &0.697  &0.978 \\
		ViT-S + iBOT & FH &6.98 &6.98  &0.704  &0.979 \\
            ViT-S + PeCLR & FH &8.51  &8.76  &0.629 &0.961  \\\midrule
		\rowcolor{Gray} ViT-S + HandMIM & FH &\textbf{6.57} &\textbf{6.57} &\textbf{0.725} &\textbf{0.984}  \\\midrule
		ViT-S & HO3Dv2 &8.71  &9.05 &0.571  &0.965 \\
		ViT-S + iBOT & HO3Dv2 &8.56 &8.84  &0.581  &0.966 \\
            ViT-S + PeCLR & HO3Dv2 & 8.81  &9.14  &0.565 &0.963  \\\midrule
		\rowcolor{Gray} ViT-S + HandMIM & HO3Dv2 &\textbf{8.22} &\textbf{8.57} &\textbf{0.597} &\textbf{0.970}  \\
		\bottomrule
	\end{tabular}
	}  \vspace{-12pt}
	\label{table:semi_cmp}
\end{table}

\begin{table}
    \centering
    \caption{\small \textbf{Cross-dataset analysis on HO3D and FreiHAND.} Methods are trained on FreiHAND and tested on HO3D, and vice versa. $^*$ indicates the methods are trained and tested on the same dataset. Performances of~\cite{hasson2019learning, hampali2020honnotate} and~\cite{spurr2021self, ziani2022tempclr} are acquired from~\cite{ hampali2020honnotate} and~\cite{ziani2022tempclr} respectively.}\vspace{3pt}
    \label{tab:cross_dataset}
    \adjustbox{width=\linewidth}{
 \setlength{\tabcolsep}{2.5pt}
    \begin{tabular}{L{85pt}C{48pt}C{48pt}C{48pt}C{48pt}}\toprule
    \multirow{2}{*}{Methods} & \multicolumn{2}{c}{Train FH/Test HO3D} & \multicolumn{2}{c}{Train HO3D/Test FH} \\ 
    \cmidrule(lr){2-3} \cmidrule(lr){4-5}
    & PAJPE$\downarrow$ & MPJPE$\downarrow$ & PAJPE$\downarrow$ & MPJPE$\downarrow$ \\ \midrule
    Hasson et al.$^*$ ~\cite{hasson2019learning}  &11.0  &31.8 &- &-\\
    Hampali et al.$^*$ ~\cite{hampali2020honnotate}  &10.7  &30.4 &-  &- \\
    PeCLR~\cite{spurr2021self}  & 13.6  & - &17.8  & -  \\
    TempCLR~\cite{ziani2022tempclr} & 13.6 & - &17.0  & -  \\\midrule
    \rowcolor{Gray} HandMIM/ViT-S &\textbf{9.9} &\textbf{30.4} &\textbf{14.1} &\textbf{29.74} \\
    \bottomrule
    \end{tabular}
    } 
\vspace{-5pt}
\end{table}

\noindent\textbf{Cross-dataset Validation.}
To evaluate the generalizability of our proposed method, we conducted a cross-data validation on 3D hand mesh estimation tasks. Specifically, we fine-tuned our model on the training set of FreiHAND and evaluated its performance on the test set of HO3D v2, and vice versa. Our results, presented in \cref{tab:cross_dataset}, demonstrate significant improvements compared to existing self-supervised methods such as PeCLR~\cite{spurr2021self} or TempCLR~\cite{ziani2022tempclr}, which indicates the superiority of our approach. Notably, our method even outperforms some recent fully supervised methods~\cite{hasson2019learning, hampali2020honnotate} when evaluated on the HO3D v2 test set.

%%%%%%%%%%%%%%%%%%%%%%%%%% Vis of Cmp FH %%%%%%%%%%%%%%%%%%%%%%%%%%
\begin{figure}[t]
	\centering
	\includegraphics[width = 0.47 \textwidth]{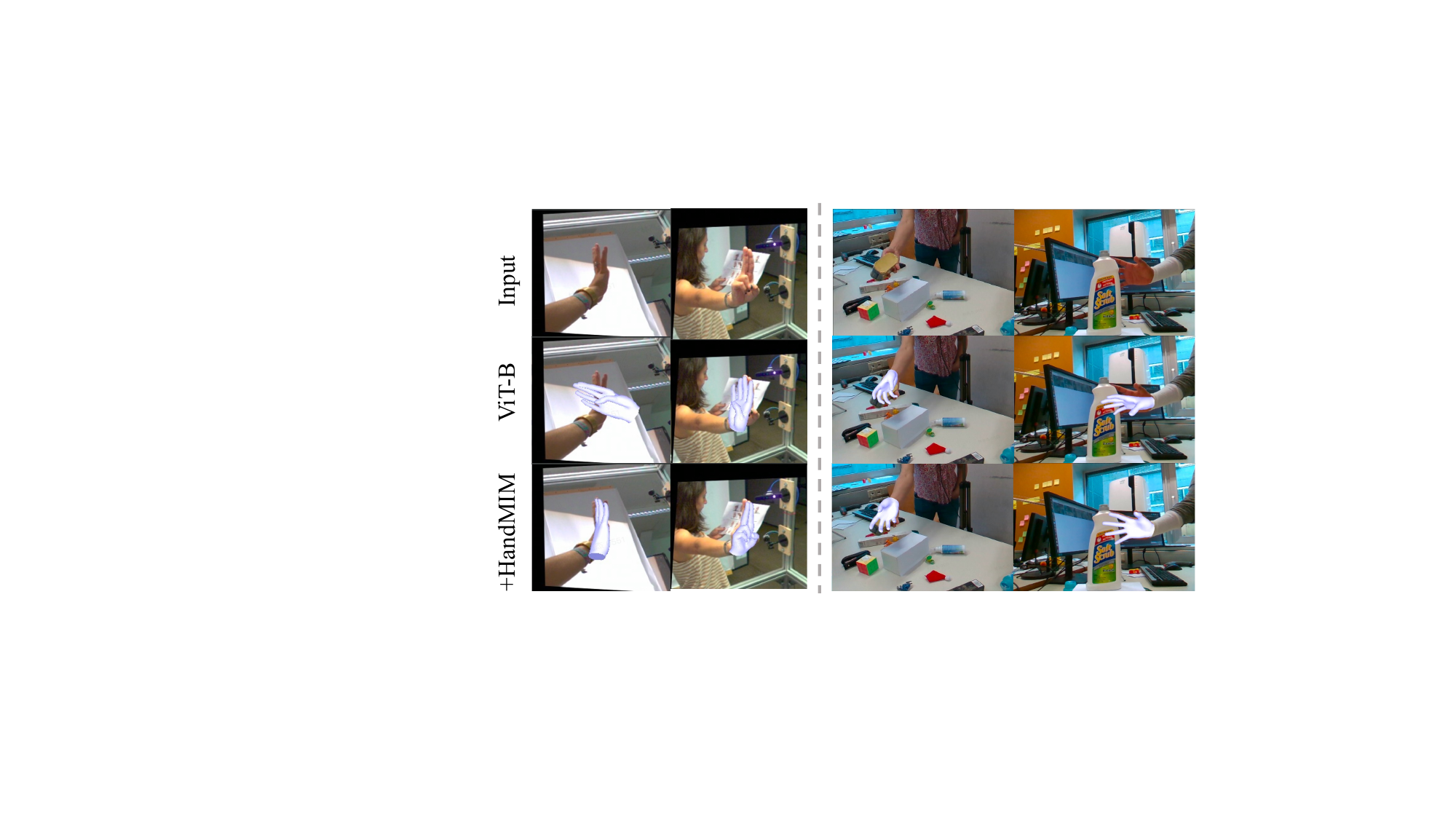}
    \caption{\small \textbf{Visualizations of HandMIM pre-training. }Left column are from FreiHAND~\cite{Freihand2019} test set while the right column are from HO3D v2~\cite{hampali2020honnotate}. Using ViT-Base as our backbone, we visualize the predicted mesh before (ViT-B) and after (+HandMIM) self-supervised training. We obtain more precise predictions after HandMIM pre-training. }
	\label{fig:cmp_vis_baseline}\vspace{-12pt}
\end{figure}

\noindent\textbf{Visualizations of HandMIM pre-training. }We are curious about the effects of hand pose estimation after self-supervised pre-training and visualize the results before and after pre-train in \cref{fig:cmp_vis_baseline}. The findings demonstrate that HandMIM pre-training enhances the resilience of 3D hand mesh estimation tasks, indicating the beneficial effects of pre-training. Specifically, the results clearly highlight the positive influence of pre-training on the robustness of hand pose estimation.

\begin{figure}[t]
    \centering
    \includegraphics[width = .48 \textwidth]{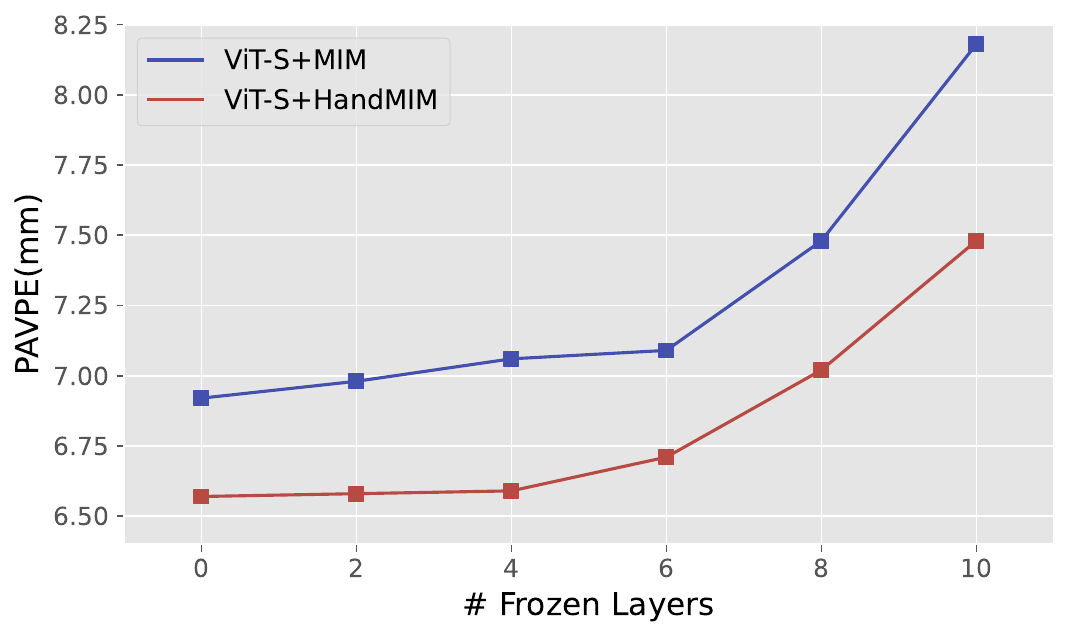}
    \caption{\small \textbf{Partial fine-tuning performance} comparison between pre-trained weight from mainstream masked image modeling methods and our HandMIM with ViT-Smallf as the backbone. We use FreiHAND~\cite{Freihand2019} test set as metric and adopt iBOT~\cite{zhou2021ibot} as baselines of MIM methods. We gradually freeze different numbers of blocks to reveal the feature generalizability learned from pre-training. }
\label{fig:prompt}\vspace{-12pt}
\end{figure}

\noindent\textbf{Partial fine-tuning. } To further explore the efficacy of the learned features, we employ a partial fine-tuning method based on the protocol proposed in~\cite{he2022masked}. We sequentially freeze the first several layers while fine-tuning the remaining transformer blocks. The results are presented in \cref{fig:prompt}, which indicate that when we freeze around half of the layers (i.e., 4 out of 12), the HandMIM approach shows only a minor decrease in accuracy compared to mainstream masked image modeling methods. Moreover, when we freeze eight or more layers, the performance gap between our method and most fully supervised methods becomes more pronounced. These findings suggest that our approach can effectively learn multi-level hand representations via our multi-level learning approach.

\begin{table}[t]
  \caption{\small \textbf{Ablation studies.} We perform ablations on the loss design of HandMIM. Specifically, we orderly remove all the three critical losses $\mathcal{L}_{\text{pose}}$, $\mathcal{L}_{\text{patch}}$, and $\mathcal{L}_{\text{recon}}$ respectively. We conduct experiments based on ViT-Small backbone and FreiHAND~\cite{Freihand2019} datasets under the same setting as the main experiments. We can conclude that every self-supervised learning target by our design is effective. } \vspace{3pt}
  \centering
  \adjustbox{width=\linewidth}{
    \begin{tabular}{C{30pt}C{30pt}C{30pt}C{35pt}C{35pt}C{35pt}C{35pt}}\toprule
    $\mathcal{L}_{\text{pose}}$ & $\mathcal{L}_{\text{patch}}$ & $\mathcal{L}_{\text{recon}}$ &PAVPE$\downarrow$ & PAJPE$\downarrow$ & F@5$\uparrow$ & F@15$\uparrow$ \\\toprule
    \rowcolor{Gray}\cmark & \cmark & \cmark & \textbf{6.57} & \textbf{6.57} & \textbf{0.725} & \textbf{0.984}  \\\midrule
    \xmark & \cmark & \cmark & 6.89 & 6.87 & 0.707 & 0.981 \\
    \cmark & \xmark & \cmark & 6.90 & 6.85 & 0.708 & 0.981 \\
    \cmark & \cmark & \xmark & 6.76 & 6.74 & 0.715 & 0.982 \\\bottomrule
    \end{tabular}%
    }
  \label{tab:ablation}\vspace{-5pt}
\end{table}%

\noindent\textbf{Ablations on self-supervised loss designs. }The pose-aware $\mathcal{L}_{\text{pose}}$, token-level $\mathcal{L}_{\text{patch}}$, and pixel-level $\mathcal{L}_{\text{recon}}$ losses in our HandMIM framework collaborate to capture distinct levels of representations from input images in a self-supervised manner. To verify the effectiveness of our design, we conduct experiments by removing one of the losses from our framework, as shown in \cref{tab:ablation}. The results demonstrate that the removal of any one of the losses results in a decrease in overall precision, highlighting the importance of our multi-level loss design. Therefore, our approach can effectively leverage diverse levels of information to enhance the robustness and accuracy of hand pose estimation tasks.

\section{Conclusion}
\label{sec:conclusion}

In this paper, we have proposed HandMIM, a unified and novel self-supervised pre-training strategy for challenging 3D hand pose estimation tasks based on mask image modeling. Unlike existing MIM pre-training methods designed for image classification, HandMIM is specifically tailored to capture hand-related knowledge, including pose and multi-level features. Extensive experiments have demonstrated that HandMIM is more relevant to hand mesh estimation tasks. We have achieved state-of-the-art in mainstream hand pose datasets FreiHAND and HO3Dv2 through a simple vanilla vision transformer backbone. In the future, we attempt to extend HandMIM to more variant structures and more human-related regression and estimation tasks. 

{\small
\bibliographystyle{ieee_fullname}
\bibliography{egbib}

\begin{thebibliography}{10}\itemsep=-1pt

\bibitem{baek2019pushing}
Seungryul Baek, Kwang~In Kim, and Tae-Kyun Kim.
\newblock Pushing the envelope for rgb-based dense 3d hand pose estimation via
  neural rendering.
\newblock In {\em CVPR}, pages 1067--1076, 2019.

\bibitem{baek2020weakly}
Seungryul Baek, Kwang~In Kim, and Tae-Kyun Kim.
\newblock Weakly-supervised domain adaptation via gan and mesh model for
  estimating 3d hand poses interacting objects.
\newblock In {\em CVPR}, pages 6121--6131, 2020.

\bibitem{bao2021beit}
Hangbo Bao, Li Dong, and Furu Wei.
\newblock Beit: Bert pre-training of image transformers.
\newblock {\em arXiv preprint arXiv:2106.08254}, 2021.

\bibitem{boukhayma20193d}
Adnane Boukhayma, Rodrigo~de Bem, and Philip~HS Torr.
\newblock 3d hand shape and pose from images in the wild.
\newblock In {\em CVPR}, pages 10843--10852, 2019.

\bibitem{cai20203d}
Yujun Cai, Liuhao Ge, Jianfei Cai, Nadia~Magnenat Thalmann, and Junsong Yuan.
\newblock 3d hand pose estimation using synthetic data and weakly labeled rgb
  images.
\newblock {\em T-PAMI}, 43(11):3739--3753, 2020.

\bibitem{cai2018weakly}
Yujun Cai, Liuhao Ge, Jianfei Cai, and Junsong Yuan.
\newblock Weakly-supervised 3d hand pose estimation from monocular rgb images.
\newblock In {\em ECCV}, pages 666--682, 2018.

\bibitem{cao2017realtime}
Zhe Cao, Tomas Simon, Shih-En Wei, and Yaser Sheikh.
\newblock Realtime multi-person 2d pose estimation using part affinity fields.
\newblock In {\em CVPR}, pages 7291--7299, 2017.

\bibitem{caron2021emerging}
Mathilde Caron, Hugo Touvron, Ishan Misra, Herv{\'e} J{\'e}gou, Julien Mairal,
  Piotr Bojanowski, and Armand Joulin.
\newblock Emerging properties in self-supervised vision transformers.
\newblock In {\em ICCV}, pages 9650--9660, 2021.

\bibitem{chen2021i2uv}
Ping Chen, Yujin Chen, Dong Yang, Fangyin Wu, Qin Li, Qingpei Xia, and Yong
  Tan.
\newblock I2uv-handnet: Image-to-uv prediction network for accurate and
  high-fidelity 3d hand mesh modeling.
\newblock In {\em ICCV}, pages 12929--12938, 2021.

\bibitem{chen2020simple}
Ting Chen, Simon Kornblith, Mohammad Norouzi, and Geoffrey Hinton.
\newblock A simple framework for contrastive learning of visual
  representations.
\newblock In {\em ICML}, pages 1597--1607. PMLR, 2020.

\bibitem{chen2020big}
Ting Chen, Simon Kornblith, Kevin Swersky, Mohammad Norouzi, and Geoffrey~E
  Hinton.
\newblock Big self-supervised models are strong semi-supervised learners.
\newblock {\em NeurIPS}, 33:22243--22255, 2020.

\bibitem{chen2020improved}
Xinlei Chen, Haoqi Fan, Ross Girshick, and Kaiming He.
\newblock Improved baselines with momentum contrastive learning.
\newblock {\em arXiv preprint arXiv:2003.04297}, 2020.

\bibitem{chen2022mobrecon}
Xingyu Chen, Yufeng Liu, Yajiao Dong, Xiong Zhang, Chongyang Ma, Yanmin Xiong,
  Yuan Zhang, and Xiaoyan Guo.
\newblock Mobrecon: Mobile-friendly hand mesh reconstruction from monocular
  image.
\newblock In {\em CVPR}, pages 20544--20554, 2022.

\bibitem{chen2022mobile}
Yinpeng Chen, Xiyang Dai, Dongdong Chen, Mengchen Liu, Xiaoyi Dong, Lu Yuan,
  and Zicheng Liu.
\newblock Mobile-former: Bridging mobilenet and transformer.
\newblock In {\em CVPR}, pages 5270--5279, 2022.

\bibitem{choi2020pose2mesh}
Hongsuk Choi, Gyeongsik Moon, and Kyoung~Mu Lee.
\newblock Pose2mesh: Graph convolutional network for 3d human pose and mesh
  recovery from a 2d human pose.
\newblock In {\em ECCV}, pages 769--787. Springer, 2020.

\bibitem{dong2021peco}
Xiaoyi Dong, Jianmin Bao, Ting Zhang, Dongdong Chen, Weiming Zhang, Lu Yuan,
  Dong Chen, Fang Wen, and Nenghai Yu.
\newblock Peco: Perceptual codebook for bert pre-training of vision
  transformers.
\newblock {\em arXiv preprint arXiv:2111.12710}, 2021.

\bibitem{dosovitskiy2020image}
Alexey Dosovitskiy, Lucas Beyer, Alexander Kolesnikov, Dirk Weissenborn,
  Xiaohua Zhai, Thomas Unterthiner, Mostafa Dehghani, Matthias Minderer, Georg
  Heigold, Sylvain Gelly, et~al.
\newblock An image is worth 16x16 words: Transformers for image recognition at
  scale.
\newblock {\em arXiv preprint arXiv:2010.11929}, 2020.

\bibitem{fan2021learning}
Zicong Fan, Adrian Spurr, Muhammed Kocabas, Siyu Tang, Michael~J Black, and
  Otmar Hilliges.
\newblock Learning to disambiguate strongly interacting hands via probabilistic
  per-pixel part segmentation.
\newblock In {\em 3DV}, pages 1--10. IEEE, 2021.

\bibitem{gao2022cyclehand}
Daiheng Gao, Xindi Zhang, Xingyu Chen, Andong Tan, Bang Zhang, Pan Pan, and
  Ping Tan.
\newblock Cyclehand: Increasing 3d pose estimation ability on in-the-wild
  monocular image through cyclic flow.
\newblock In {\em ACM MM}, pages 2452--2463, 2022.

\bibitem{ge20193d}
Liuhao Ge, Zhou Ren, Yuncheng Li, Zehao Xue, Yingying Wang, Jianfei Cai, and
  Junsong Yuan.
\newblock 3d hand shape and pose estimation from a single rgb image.
\newblock In {\em CVPR}, pages 10833--10842, 2019.

\bibitem{hampali2020honnotate}
Shreyas Hampali, Mahdi Rad, Markus Oberweger, and Vincent Lepetit.
\newblock Honnotate: A method for 3d annotation of hand and object poses.
\newblock In {\em CVPR}, 2020.

\bibitem{hampali2021handsformer}
Shreyas Hampali, Sayan~Deb Sarkar, Mahdi Rad, and Vincent Lepetit.
\newblock Handsformer: Keypoint transformer for monocular 3d pose estimation
  ofhands and object in interaction.
\newblock {\em arXiv preprint arXiv:2104.14639}, 2021.

\bibitem{hampali2022keypoint}
Shreyas Hampali, Sayan~Deb Sarkar, Mahdi Rad, and Vincent Lepetit.
\newblock Keypoint transformer: Solving joint identification in challenging
  hands and object interactions for accurate 3d pose estimation.
\newblock In {\em CVPR}, pages 11090--11100, 2022.

\bibitem{hasson2020leveraging}
Yana Hasson, Bugra Tekin, Federica Bogo, Ivan Laptev, Marc Pollefeys, and
  Cordelia Schmid.
\newblock Leveraging photometric consistency over time for sparsely supervised
  hand-object reconstruction.
\newblock In {\em CVPR}, pages 571--580, 2020.

\bibitem{hasson2019learning}
Yana Hasson, Gul Varol, Dimitrios Tzionas, Igor Kalevatykh, Michael~J Black,
  Ivan Laptev, and Cordelia Schmid.
\newblock Learning joint reconstruction of hands and manipulated objects.
\newblock In {\em CVPR}, pages 11807--11816, 2019.

\bibitem{he2022masked}
Kaiming He, Xinlei Chen, Saining Xie, Yanghao Li, Piotr Doll{\'a}r, and Ross
  Girshick.
\newblock Masked autoencoders are scalable vision learners.
\newblock In {\em CVPR}, pages 16000--16009, 2022.

\bibitem{jin2020whole}
Sheng Jin, Lumin Xu, Jin Xu, Can Wang, Wentao Liu, Chen Qian, Wanli Ouyang, and
  Ping Luo.
\newblock Whole-body human pose estimation in the wild.
\newblock In {\em ECCV}, pages 196--214. Springer, 2020.

\bibitem{kingma2014adam}
Diederik~P Kingma and Jimmy Ba.
\newblock Adam: A method for stochastic optimization.
\newblock {\em arXiv preprint arXiv:1412.6980}, 2014.

\bibitem{kulon2020weakly}
Dominik Kulon, Riza~Alp Guler, Iasonas Kokkinos, Michael~M Bronstein, and
  Stefanos Zafeiriou.
\newblock Weakly-supervised mesh-convolutional hand reconstruction in the wild.
\newblock In {\em CVPR}, pages 4990--5000, 2020.

\bibitem{li2021artiboost}
Kailin Li, Lixin Yang, Xinyu Zhan, Jun Lv, Wenqiang Xu, Jiefeng Li, and Cewu
  Lu.
\newblock Artiboost: Boosting articulated 3d hand-object pose estimation via
  online exploration and synthesis.
\newblock {\em arXiv preprint arXiv:2109.05488}, 2021.

\bibitem{li2022interacting}
Mengcheng Li, Liang An, Hongwen Zhang, Lianpeng Wu, Feng Chen, Tao Yu, and
  Yebin Liu.
\newblock Interacting attention graph for single image two-hand reconstruction.
\newblock In {\em CVPR}, pages 2761--2770, 2022.

\bibitem{lin2021end}
Kevin Lin, Lijuan Wang, and Zicheng Liu.
\newblock End-to-end human pose and mesh reconstruction with transformers.
\newblock In {\em CVPR}, pages 1954--1963, 2021.

\bibitem{lin2021mesh}
Kevin Lin, Lijuan Wang, and Zicheng Liu.
\newblock Mesh graphormer.
\newblock In {\em ICCV}, pages 12939--12948, 2021.

\bibitem{liu2021semi}
Shaowei Liu, Hanwen Jiang, Jiarui Xu, Sifei Liu, and Xiaolong Wang.
\newblock Semi-supervised 3d hand-object poses estimation with interactions in
  time.
\newblock In {\em CVPR}, 2021.

\bibitem{loshchilov2017decoupled}
Ilya Loshchilov and Frank Hutter.
\newblock Decoupled weight decay regularization.
\newblock {\em arXiv preprint arXiv:1711.05101}, 2017.

\bibitem{lugaresi2019mediapipe}
Camillo Lugaresi, Jiuqiang Tang, Hadon Nash, Chris McClanahan, Esha Uboweja,
  Michael Hays, Fan Zhang, Chuo-Ling Chang, Ming~Guang Yong, Juhyun Lee, et~al.
\newblock Mediapipe: A framework for building perception pipelines.
\newblock {\em arXiv preprint arXiv:1906.08172}, 2019.

\bibitem{mehta2021mobilevit}
Sachin Mehta and Mohammad Rastegari.
\newblock Mobilevit: light-weight, general-purpose, and mobile-friendly vision
  transformer.
\newblock {\em arXiv preprint arXiv:2110.02178}, 2021.

\bibitem{Moon_2020_ECCV_I2L-MeshNet}
Gyeongsik Moon and Kyoung~Mu Lee.
\newblock I2l-meshnet: Image-to-lixel prediction network for accurate 3d human
  pose and mesh estimation from a single rgb image.
\newblock In {\em ECCV}, 2020.

\bibitem{moon2020interhand2}
Gyeongsik Moon, Shoou-I Yu, He Wen, Takaaki Shiratori, and Kyoung~Mu Lee.
\newblock Interhand2. 6m: A dataset and baseline for 3d interacting hand pose
  estimation from a single rgb image.
\newblock In {\em ECCV}, pages 548--564. Springer, 2020.

\bibitem{oord2018representation}
Aaron van~den Oord, Yazhe Li, and Oriol Vinyals.
\newblock Representation learning with contrastive predictive coding.
\newblock {\em arXiv preprint arXiv:1807.03748}, 2018.

\bibitem{pan2022edgevits}
Junting Pan, Adrian Bulat, Fuwen Tan, Xiatian Zhu, Lukasz Dudziak, Hongsheng
  Li, Georgios Tzimiropoulos, and Brais Martinez.
\newblock Edgevits: Competing light-weight cnns on mobile devices with vision
  transformers.
\newblock In {\em ECCV}, pages 294--311. Springer, 2022.

\bibitem{Park_2022_CVPR_HandOccNet}
JoonKyu Park, Yeonguk Oh, Gyeongsik Moon, Hongsuk Choi, and Kyoung~Mu Lee.
\newblock Handoccnet: Occlusion-robust 3d hand mesh estimation network.
\newblock In {\em CVPR}, 2022.

\bibitem{rao2021dynamicvit}
Yongming Rao, Wenliang Zhao, Benlin Liu, Jiwen Lu, Jie Zhou, and Cho-Jui Hsieh.
\newblock Dynamicvit: Efficient vision transformers with dynamic token
  sparsification.
\newblock {\em NeurIPS}, 34:13937--13949, 2021.

\bibitem{romero2022embodied}
Javier Romero, Dimitrios Tzionas, and Michael~J Black.
\newblock Embodied hands: Modeling and capturing hands and bodies together.
\newblock {\em arXiv preprint arXiv:2201.02610}, 2022.

\bibitem{spurr2021self}
Adrian Spurr, Aneesh Dahiya, Xi Wang, Xucong Zhang, and Otmar Hilliges.
\newblock Self-supervised 3d hand pose estimation from monocular rgb via
  contrastive learning.
\newblock In {\em ICCV}, pages 11230--11239, 2021.

\bibitem{spurr2020weakly}
Adrian Spurr, Umar Iqbal, Pavlo Molchanov, Otmar Hilliges, and Jan Kautz.
\newblock Weakly supervised 3d hand pose estimation via biomechanical
  constraints.
\newblock In {\em ECCV}, pages 211--228. Springer, 2020.

\bibitem{tang2021towards}
Xiao Tang, Tianyu Wang, and Chi-Wing Fu.
\newblock Towards accurate alignment in real-time 3d hand-mesh reconstruction.
\newblock In {\em ICCV}, pages 11698--11707, 2021.

\bibitem{tian2020contrastive}
Yonglong Tian, Dilip Krishnan, and Phillip Isola.
\newblock Contrastive multiview coding.
\newblock In {\em ECCV}, pages 776--794. Springer, 2020.

\bibitem{tse2022s}
Tze Ho~Elden Tse, Zhongqun Zhang, Kwang~In Kim, Ales Leonardis, Feng Zheng, and
  Hyung~Jin Chang.
\newblock S$^2$contact: Graph-based network for 3d hand-object contact
  estimation with semi-supervised learning.
\newblock In {\em ECCV}, pages 568--584. Springer, 2022.

\bibitem{wei2022masked}
Chen Wei, Haoqi Fan, Saining Xie, Chao-Yuan Wu, Alan Yuille, and Christoph
  Feichtenhofer.
\newblock Masked feature prediction for self-supervised visual pre-training.
\newblock In {\em CVPR}, pages 14668--14678, 2022.

\bibitem{xie2022simmim}
Zhenda Xie, Zheng Zhang, Yue Cao, Yutong Lin, Jianmin Bao, Zhuliang Yao, Qi
  Dai, and Han Hu.
\newblock Simmim: A simple framework for masked image modeling.
\newblock In {\em CVPR}, pages 9653--9663, 2022.

\bibitem{yang2021cpf}
Lixin Yang, Xinyu Zhan, Kailin Li, Wenqiang Xu, Jiefeng Li, and Cewu Lu.
\newblock Cpf: Learning a contact potential field to model the hand-object
  interaction.
\newblock In {\em ICCV}, pages 11097--11106, 2021.

\bibitem{zeng2022not}
Wang Zeng, Sheng Jin, Wentao Liu, Chen Qian, Ping Luo, Wanli Ouyang, and
  Xiaogang Wang.
\newblock Not all tokens are equal: Human-centric visual analysis via token
  clustering transformer.
\newblock In {\em CVPR}, pages 11101--11111, 2022.

\bibitem{pymaf2021}
Hongwen Zhang, Yating Tian, Xinchi Zhou, Wanli Ouyang, Yebin Liu, Limin Wang,
  and Zhenan Sun.
\newblock Pymaf: 3d human pose and shape regression with pyramidal mesh
  alignment feedback loop.
\newblock In {\em ICCV}, 2021.

\bibitem{zhang2021hand}
Xiong Zhang, Hongsheng Huang, Jianchao Tan, Hongmin Xu, Cheng Yang, Guozhu
  Peng, Lei Wang, and Ji Liu.
\newblock Hand image understanding via deep multi-task learning.
\newblock In {\em ICCV}, pages 11281--11292, 2021.

\bibitem{zhang2019end}
Xiong Zhang, Qiang Li, Hong Mo, Wenbo Zhang, and Wen Zheng.
\newblock End-to-end hand mesh recovery from a monocular rgb image.
\newblock In {\em ICCV}, pages 2354--2364, 2019.

\bibitem{zhou2021ibot}
Jinghao Zhou, Chen Wei, Huiyu Wang, Wei Shen, Cihang Xie, Alan Yuille, and Tao
  Kong.
\newblock ibot: Image bert pre-training with online tokenizer.
\newblock {\em ICLR}, 2022.

\bibitem{ziani2022tempclr}
Andrea Ziani, Zicong Fan, Muhammed Kocabas, Sammy Christen, and Otmar Hilliges.
\newblock Tempclr: Reconstructing hands via time-coherent contrastive learning.
\newblock {\em arXiv preprint arXiv:2209.00489}, 2022.

\bibitem{zimmermann2021contrastive}
Christian Zimmermann, Max Argus, and Thomas Brox.
\newblock Contrastive representation learning for hand shape estimation.
\newblock In {\em DAGM German Conference on Pattern Recognition}, pages
  250--264. Springer, 2021.

\bibitem{Freihand2019}
Christian Zimmermann, Duygu Ceylan, Jimei Yang, Bryan Russell, Max Argus, and
  Thomas Brox.
\newblock Freihand: A dataset for markerless capture of hand pose and shape
  from single rgb images.
\newblock In {\em ICCV}, pages 813--822, 2019.

\end{thebibliography}
}

\end{document}